\documentclass[conference]{IEEEtran}
\IEEEoverridecommandlockouts
\usepackage{cite}
\usepackage{amsmath,amssymb,amsfonts}
\usepackage{graphicx}
\usepackage{textcomp}
\usepackage{xcolor}

\usepackage{algorithm}
\usepackage{algpseudocode}

\usepackage{xcolor}

\usepackage{tabularx,booktabs}
\usepackage[flushleft]{threeparttable}

\def\BibTeX{{\rm B\kern-.05em{\sc i\kern-.025em b}\kern-.08em
    T\kern-.1667em\lower.7ex\hbox{E}\kern-.125emX}}
    
\usepackage{relsize}
\usepackage{scalerel}
\usepackage{tikz}
\usetikzlibrary{svg.path}

\definecolor{orcidlogocol}{HTML}{A6CE39}
\tikzset{
    orcidlogo/.pic={
        \fill[orcidlogocol] svg{M256,128c0,70.7-57.3,128-128,128C57.3,256,0,198.7,0,128C0,57.3,57.3,0,128,0C198.7,0,256,57.3,256,128z};
        \fill[white] svg{M86.3,186.2H70.9V79.1h15.4v48.4V186.2z}
        svg{M108.9,79.1h41.6c39.6,0,57,28.3,57,53.6c0,27.5-21.5,53.6-56.8,53.6h-41.8V79.1z M124.3,172.4h24.5c34.9,0,42.9-26.5,42.9-39.7c0-21.5-13.7-39.7-43.7-39.7h-23.7V172.4z}
        svg{M88.7,56.8c0,5.5-4.5,10.1-10.1,10.1c-5.6,0-10.1-4.6-10.1-10.1c0-5.6,4.5-10.1,10.1-10.1C84.2,46.7,88.7,51.3,88.7,56.8z};
    }
}

\newcommand\orcidicon[1]{\href{https://orcid.org/#1}{\mbox{\scalerel*{
                \begin{tikzpicture}[yscale=-1,transform shape]
                \pic{orcidlogo};
                \end{tikzpicture}
            }{|}}}}

\usepackage[colorlinks = true,
            linkcolor = blue,
            urlcolor  = magenta,
            citecolor = green,
            anchorcolor = blue]{hyperref} 
    
\begin{document}


\title{POCS-based Clustering Algorithm\\
\thanks{*Corresponding Author}
}


\author{\IEEEauthorblockN{Le-Anh Tran$^{\textsuperscript{\orcidicon{0000-0002-9380-7166}}}$}
\IEEEauthorblockA{\textit{Dept. of Electronics Engineering} \\
\textit{Myongji University}\\
Gyeonggi, South Korea \\
leanhtran@mju.ac.kr}
\and
\IEEEauthorblockN{Henock M. Deberneh}
\IEEEauthorblockA{\textit{Dept. of Biochemistry and Molecular Biology} \\
\textit{University of Texas Medical Branch}\\
Texas, United States \\
henockmamo54@gmail.com}
\and
\IEEEauthorblockN{Truong-Dong Do$^{\textsuperscript{\orcidicon{0000-0001-8178-0018}}}$}
\IEEEauthorblockA{\textit{Dept. of Aerospace Engineering} \\
\textit{Sejong University}\\
Seoul, South Korea \\
dongdo@sju.ac.kr}
\and
\IEEEauthorblockN{Thanh-Dat Nguyen$^{\textsuperscript{\orcidicon{0000-0002-7098-5816}}}$}
\IEEEauthorblockA{\textit{Dept. of Research and Development} \\
\textit{OCST Co., Ltd.} \\
Seoul, South Korea \\
thanhdat6716@gmail.com}
\and
\IEEEauthorblockN{My-Ha Le}
\IEEEauthorblockA{\textit{Dept. of Electrical and Electronics Engineering} \\
\textit{HCMC University of Technology and Education}\\
Ho Chi Minh City, Vietnam \\
halm@hcmute.edu.vn }
\and
\IEEEauthorblockN{Dong-Chul Park*}
\IEEEauthorblockA{\textit{Dept. of Electronics Engineering} \\
\textit{Myongji University}\\
Gyeonggi, South Korea \\
parkd@mju.ac.kr}

}

\maketitle

\begin{abstract}
A novel clustering technique based on the projection onto convex set (POCS) method, called POCS-based clustering algorithm, is proposed in this paper. The proposed POCS-based clustering algorithm exploits a parallel projection method of POCS to find appropriate cluster prototypes in the feature space. The algorithm considers each data point as a convex set and projects the cluster prototypes parallelly to the member data points. The projections are convexly combined to minimize the objective function for data clustering purpose. The performance of the proposed POCS-based clustering algorithm is verified through experiments on various synthetic datasets. The experimental results show that the proposed POCS-based clustering algorithm is competitive and efficient in terms of clustering error and execution speed when compared with other conventional clustering methods including Fuzzy C-Means (FCM) and K-Means clustering algorithms. Code is available at: \url{https://github.com/tranleanh/pocs-based-clustering} 
\end{abstract}


\section{Introduction}

Projection onto convex set (POCS) is a powerful tool for signal synthesis and image restoration which was originally introduced by Bregman in the mid-1960s \cite{b1}. The POCS method has been widely used to find a common point of convex sets in several signal processing problems. The main target of the POCS approach is to find a vector that resides in the intersection of convex sets. Bregman has shown that successive projections between two or more convex sets with non-empty intersection converge to a point that exists in the intersection of the convex sets. In the case of disjoint closed convex sets, the sequential projection does not converge to a single point, instead it converges to greedy limit cycles which are dependent on the order of the projections \cite{b1}. This property of POCS, however, can be applied to clustering problems. 

Clustering is an unsupervised data analysis technique that categories similar data points while separating them from the different ones \cite{b2}. Most clustering algorithms try to find homogeneous subgroups that have similar characteristics by the type of metric employed. The K-Means clustering algorithm, which has been one of the most popular methods for general clustering purposes \cite{b9}, uses the Euclidean distance to measure the similarity \cite{b2}. The K-Means clustering algorithm alternates between assigning cluster membership for each data point to the nearest cluster center and computing the center of each cluster as the prototype of its member data points. The objective of the K-Means clustering algorithm is to find a set of prototypes that minimize the cost function. The K-Means clustering algorithm terminates its training procedure when there is no further change in the assignment of instances to clusters \cite{b2}. The convergence of the K-Means clustering algorithm heavily depends on the initial prototypes. However, there exists no efficient and universal method for identifying the initial partitions \cite{b3}. Furthermore, the K-Means algorithm is known to be sensitive to noise and outliers \cite{b2}. In the Fuzzy C-Means (FCM) clustering algorithm \cite{b4}, on the other hand, a data point can belong to multiple subgroups simultaneously. The degree of certainty for a data point belonging to a certain cluster is represented by a membership function. The performance of the FCM algorithm is highly dependent on the selection of the initial prototypes and the initial membership value \cite{b4}. Furthermore, the drawbacks of the FCM clustering algorithm include extended computational time, incapability in handling noisy data and outliers \cite{b4}. In order to improve the convergence speed and the computation complexity of the FCM algorithm, the Gradient-Based Fuzzy C-Means (GBFCM) algorithm \cite{b5} was introduced by Park and Dagher which combines FCM and the characteristics of Kohonen’s Self Organizing Map \cite{b6} to improve performance.

In this paper, we propose a novel clustering algorithm using the convergence property of POCS. The proposed POCS-based clustering algorithm considers each data point as a convex set and projects the prototypes of the clusters to each of its constituent instances to compute a new set of center points. At first, the proposed algorithm initializes \textit{k} cluster prototypes. Based on the distance to the prototypes, each data point is assigned to one of the clusters which have the minimum distance from the data point. The cluster prototypes are projected to the member data points and combined convexly to minimize the objective function and the algorithm computes a new set of prototypes.

The remainder of this paper is structured as follows. Section II briefly reviews the POCS method. POCS-based clustering algorithm is proposed in Section III. In Section IV, the performance of the proposed POCS-based clustering algorithm on various synthetic datasets is examined and compared with those of other conventional clustering methods. Finally, Section V concludes the paper.

\section{The POCS Method}

\subsection{Convex Set}

\begin{figure}[t]
\centering
\includegraphics[width=0.7\columnwidth]{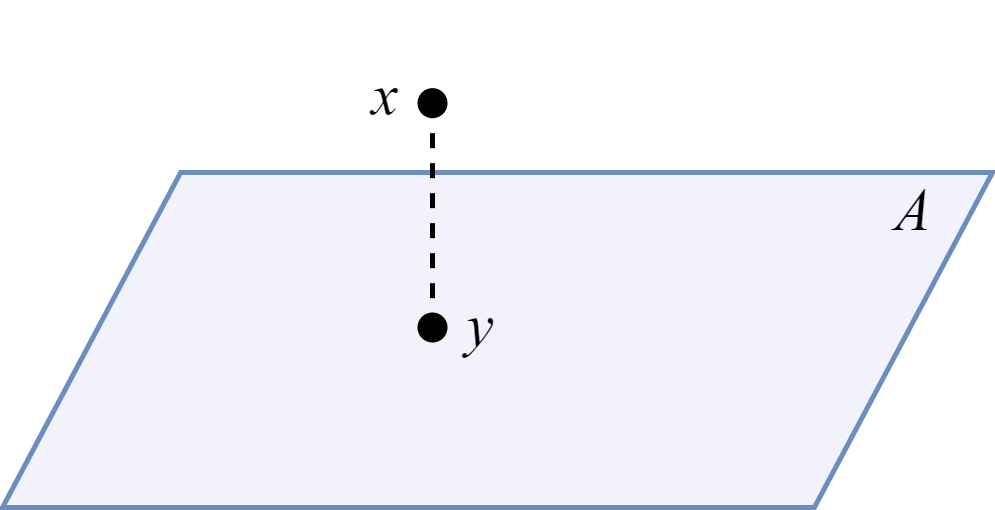}
\caption{Projection onto convex set: the projection of $x$ onto $A$ is the unique element in $A$ which is closest to $x$ and is denoted as $y$.}
\end{figure}

The theory of convex set has a rich history and has been a focus of research. It has been one of the most powerful tools in the theory of optimization \cite{b1}. A convex set is a collection of data points having the following property: given a non-empty set $A$ which is the subset of a Hilbert space $H$, $ A \subseteq H $ is called convex, for  $ \forall x_1, x_2 \in A $ and $\forall \lambda \in [0, 1]$, if the following holds true:

\begin{equation}
x := \lambda x_1 + (1 - \lambda)x_2 \in A  \label{eq1}
\end{equation}

Note that if $\lambda = 1$, $x = x_1$, and if $\lambda = 0$, $x = x_2$. For any value of $0 \leq \lambda \leq 1$ and $ x \in A $, $x$ lies on the line segment joining $x_1$ and $x_2$ when the set is convex.

\subsection{Projection onto Convex Set}
The concept of projection of a point to a plane deals with the optimization problem of interest, which is finding a point on the plane that has a minimum distance from the center of projection. For a given point $ x \notin A $, the projection of $x$ onto $A$ is the unique point $ y \in A $ such that the distance between $x$ and $y$ is a minimum. If $ x \in A $, then the projection of $x$ onto $A$ is $x$. The constrained optimization task can be expressed as:

\begin{equation}
y = argmin || x - y^* ||^2  \label{eq2}
\end{equation} where $y^*$ is all the points on the set $A$. The projection onto a convex set is illustrated in Fig. 1.

\subsection{Alternating Projection onto Convex Sets}

\begin{figure}[t]
\centering
\includegraphics[width=1.0\columnwidth]{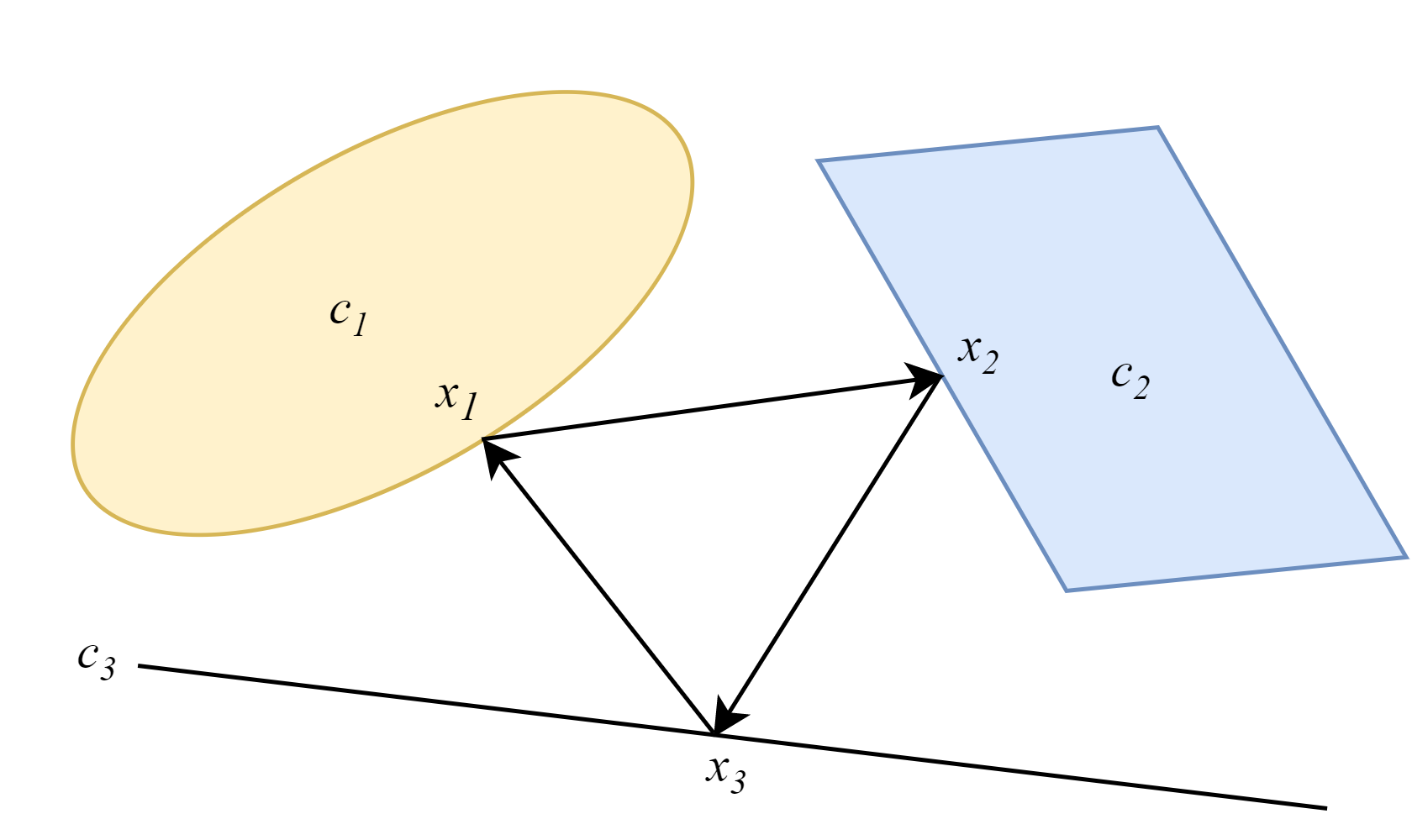}
\caption{Alternating POCS converges to a limit cycle for disjoint convex sets.}
\end{figure}

Alternating projection between two or more convex sets with non-empty intersection converges to a point that resides in the intersection of the convex sets. This prominent property of POCS can be applied to solve many optimization tasks, which can be described under the convex restriction sets. When $c_i$, $1 \leq i \leq n$, represents $n$ constraints with a non-empty intersection, the solution to the task resides in the intersection of the convex sets, which is expressed as: 


\begin{equation}
c_0 = \bigcap_{i=1}^{n}c_{i}  \label{eq3}
\end{equation}

Given the convex sets $c_i$, $ 1  \leq i \leq n$, which are closed and convex with a non-empty intersection, the successive projections on the sets will converge to a point that belongs to the intersection. Equation \eqref{eq4} denotes the algorithm, where $x_0$ is any point and represents the starting point, and $P_c$ is a projection operator onto $c$.

\begin{equation}
x_{k+1} = P_{c_n}...P_{c_2}P_{c_1}x_k \label{eq4}
\end{equation}

When these convex sets are disjoint, the sequential projection does not converge to a single point. Instead, it converges to greedy limit cycles which are dependent on the order of the projections. Fig. 2 depicts a geometrical visualization of the alternating POCS for three disjoint convex sets.

\subsection{Parallel Projection onto Convex Sets}

\begin{figure}[t]
\centering
\includegraphics[width=0.9\columnwidth]{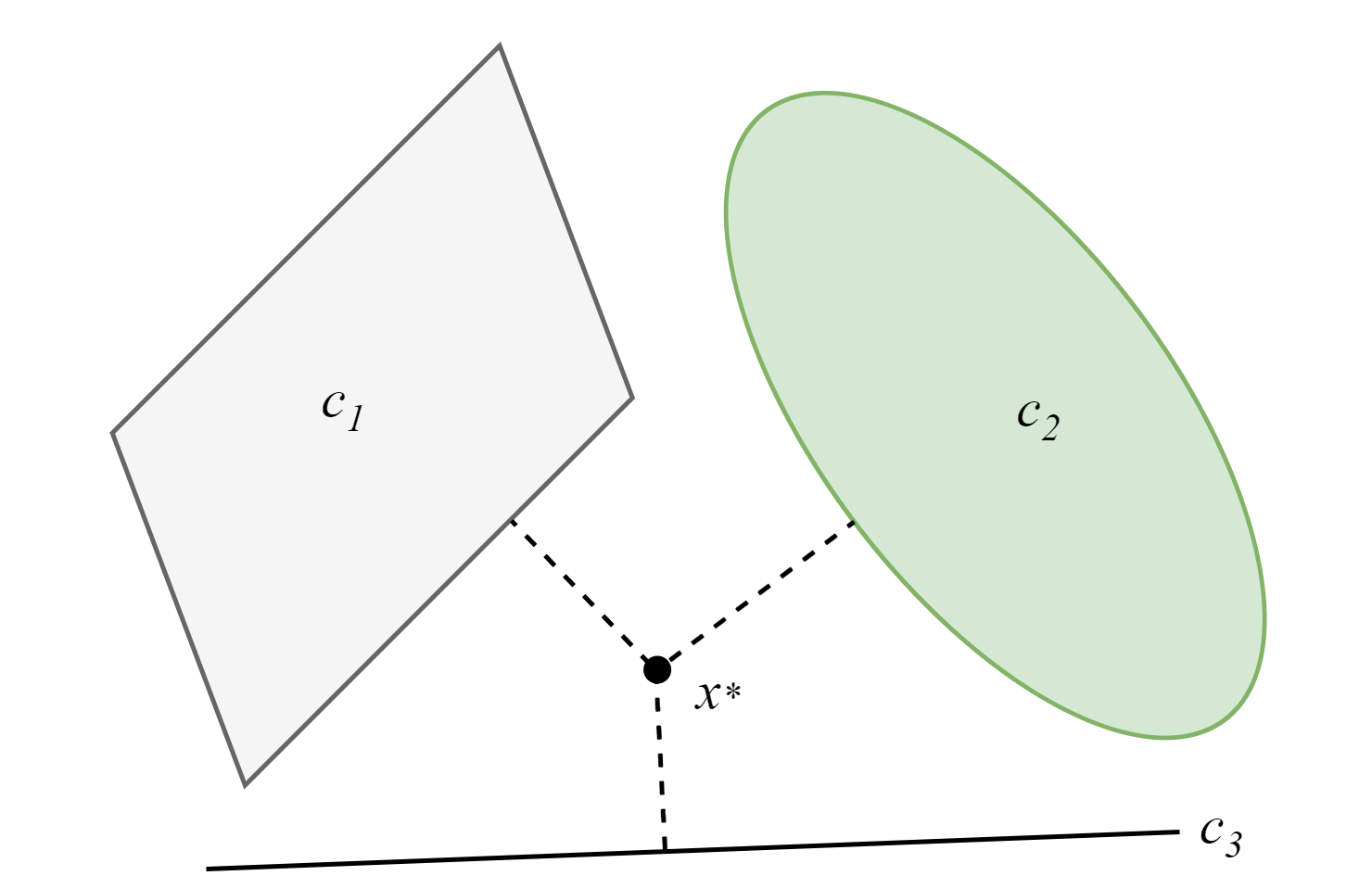}
\caption{Graphical interpretation of parallel POCS for disjoint convex sets.}
\end{figure}

In the parallel mode of POCS, the initial point is projected to all convex sets simultaneously. Each projection has a weight and is combined convexly to solve the minimization problem. For a set of $n$ convex sets $C = \{ c_i | 1 \leq i \leq n \}$, the weighted simultaneous projections can be computed as follows:

\begin{equation}
x_{k+1} = x_k + \sum_{i=1}^{n} w_i (P_{c_i} - x_k) , k=0,1,2,... \label{eq5}
\end{equation} 

\begin{equation}
\sum_{i=1}^{n} w_i = 1 \label{eq6}
\end{equation} where $P_{c_i}$ is the projection of $x_k$ onto convex set $c_i$ and $w_i$ is the weight of importance of the projection. Note that $x_k$ represents the $k^{th}$ projection of the initial point $x_0$. The projection continues until convergence. The main advantages of the parallel mode of POCS when compared with the alternating one include computational efficiency and improved execution time.

If the sets are disjoint convex sets, the parallel form of POCS converges to a point that minimizes the weighted sum of the squares of distances to the sets. Suppose that the projection converges to a point $x^{*}$ such that the distance $d$ defined by \eqref{eq7} is minimized. A graphical illustration of the convergence of the parallel POCS method is presented in Fig. 3.

\begin{equation}
d = \sum_{i=1}^{n} w_i || x^{*} -  P_{c_i}(x^{*}) ||^2   \label{eq7}
\end{equation}

\section{POCS-based Clustering Algorithm}

As mentioned in the previous section, the iterative projections (alternating or parallel) onto convex sets with non-empty intersection weakly converges to a point that resides on the intersection of the sets. For disjoint sets, the alternating POCS converges to a greedy limit cycle, the parallel mode of projection converges to a point that minimizes the weighted sum of the squared distances. In this study, we propose a clustering algorithm that utilizes the parallel form of POCS. The proposed POCS-based clustering algorithm considers each data point as a convex set and all data points in the cluster as disjoint convex sets. The objective function of the proposed POCS-based clustering algorithm is defined as:


\begin{equation}
J = argmin \sum_{j}^{k} \sum_{i=1}^{n} w_i || x_j -  P_{c_i}(x_j) ||^2   \label{eq11}
\end{equation}

\begin{equation}
w_i = \frac{|| x_j - d_i ||}{\sum_{p=1}^{n}  || x_j - d_{p} || }   \label{eq9}
\end{equation} with a constraint

\begin{equation}
\sum_{i=1}^{n} w_i = 1  \label{eq10}
\end{equation} where $k$, $n$ represents the number of clusters and the number of data points in one cluster, respectively, while $P_{c_i} (x_j)$ is the projection of the cluster prototype $x_j$ onto the member point $d_i$ and $w_i$ denotes the weight of importance of the projection.

At first, the algorithm initializes cluster prototypes as in K-Means++ \cite{b7} and assigns each data point to the nearest cluster center. Until convergence, the algorithm computes new cluster prototypes using \eqref{eq12} with a constraint as in \eqref{eq13}. The simultaneous  projections of the prototype $x_k$, where $k$ is the iteration index, continue until convergence. Starting from an initial point $x_0$, the projections converge to a point, $x_\infty$, that can minimize the weighted sum of the squares of distances.

\begin{algorithm}[t]
\caption{POCS-based Clustering Algorithm}
\begin{algorithmic}[1]
\State Initialize cluster prototypes $x_{k,0} (k = 1, 2, ..., K)$,
\State Assign each data $d$ to its closest prototype,
\State $n \gets 1$,
\While{$n < N$}

    \For {$k = 1$ to $K$}
    \State $x_{k,n} \gets x_{k,n-1}$
        \For {$i = 1$ to $I_k$}
        
            \State $w_i \gets \mathlarger{\frac{|| x_{k,n-1} - d_i ||}{\sum_{j=1}^{I_k}  || x_{k,n-1} - d_j || }}$
            
            \State $x_{k,n} \gets x_{k,n} + w_i(d_i - x_{k,n-1})$
            
        \EndFor
    \EndFor

    \If {$x_{k,n} == x_{k,n-1}, \forall k$}
        \State break     \Comment{converged!}
    \EndIf

\EndWhile
\end{algorithmic}
\end{algorithm}

\begin{equation}
x_{k+1} = x_k + \sum_{i=1}^{n} w_i (P_{c_i} - x_k), k=0,1,2,... \label{eq12}
\end{equation}

\begin{equation}
\sum_{i=1}^{n} w_i = 1  \label{eq13}
\end{equation}


\section{Experiments and Results}

\begin{table}[t]
\caption{Synthetic datasets.}
\setlength\tabcolsep{0pt} 
\begin{tabular*}{\columnwidth}{@{\extracolsep{\fill}} lccc}

\toprule
     \textbf{Dataset} & \textbf{Number of Clusters} & \textbf{Attributes} & \textbf{Instances} \\

\midrule
     A1  & 20 & 2 & 3,000 \\
     A2  & 35 & 2 & 5,250 \\
     S1  & 15 & 2 & 5,000 \\
     S2  & 15 & 2 & 5,000 \\
     R15  & 15 & 2 & 600 \\
     Aggregation  & 7 & 2 & 788 \\
     
\bottomrule
\end{tabular*}
\end{table}

In order to evaluate the effectiveness of the proposed POCS-based clustering algorithm, various experiments on a variety of synthetic datasets have been conducted. The experiments exploits publicly available synthetic datasets that are available on the website “Clustering datasets” \cite{b8}. These experiments aim to thoroughly explain the convergence property of the proposed algorithm in terms of visual clustering results,
execution speed, and clustering error. The specifications of the datasets are summarized in Table I.

\begin{figure*}[t]
    \centering
    \includegraphics[width=0.9\textwidth]{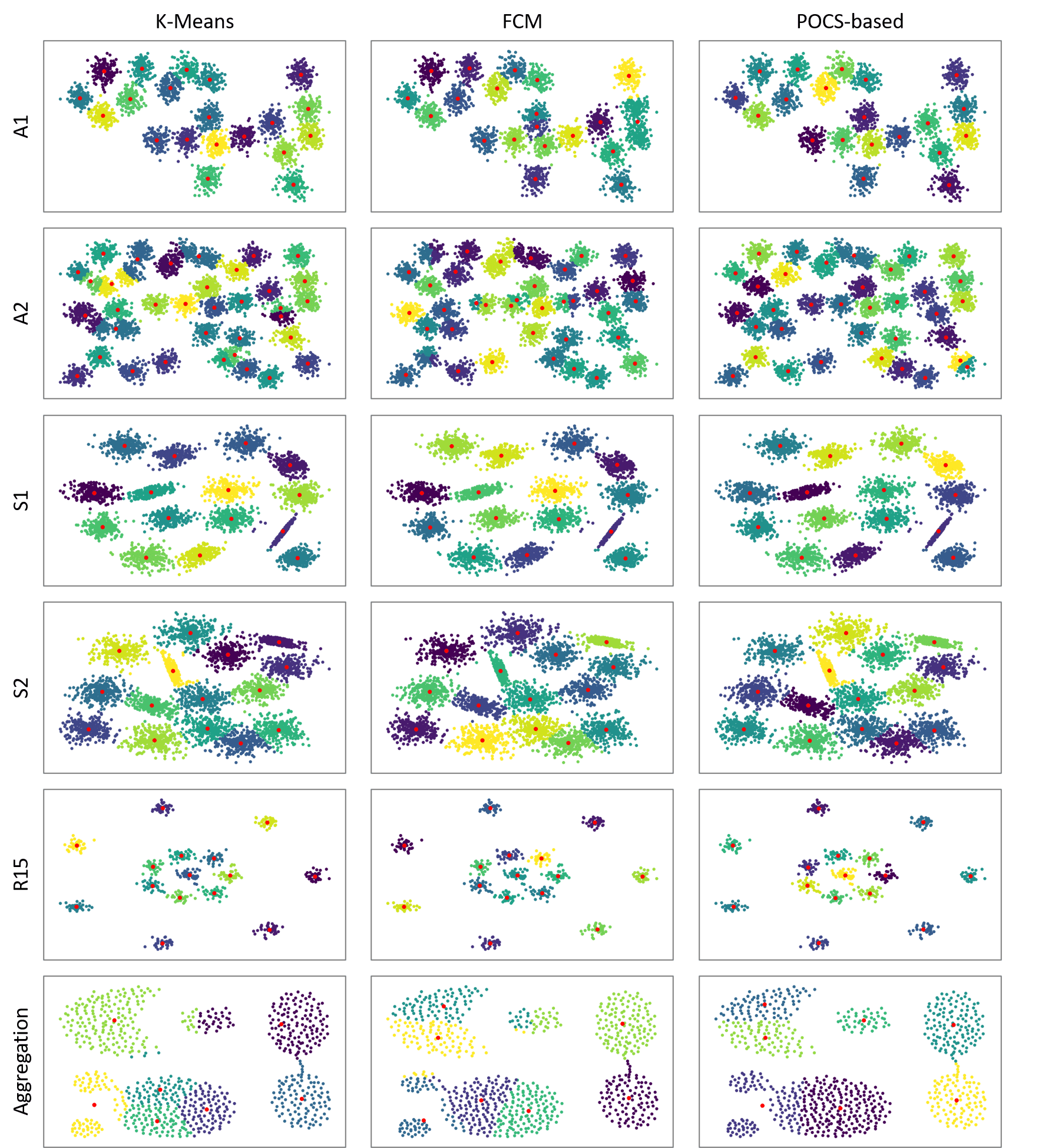}
    \caption{Clustering results of different algorithms on synthetic datasets.}
\end{figure*}

Fig. 4 illustrates the visual clustering results in two-dimensional plots where each unique color in a plot denotes a cluster obtained after convergence. Each cluster center is marked by red color and located in the vicinity of the cluster. Generally, the proposed POCS-based clustering algorithm has a competitive performance when compared against popular clustering techniques like K-Means and FCM algorithms. 

On A1 and A2 datasets which include 3,000 and 5,250 two-dimensional data points with 20 and 35 clusters, respectively, all three clustering algorithms are able to positively identify the clusters despite the existing mild overlapping among those clusters. However, the cluster shapes and the final prototypes vary in different algorithms. For S1 and S2 datasets (each dataset has 5,000 data points which are distributed to 15 clusters), the algorithms are able to pick the cluster groups with favorable results.

R15 dataset contains 600 data points which are divided into 15 clusters. One of the clusters is located in the vicinity of the center of the dataset and the remaining clusters surround the center cluster on two layers of circular orientation. As can be seen from Fig. 4, the algorithms can adequately determine the cluster prototypes and groups for R15 dataset.

On Aggregation dataset which is comprised of 7 clusters with a total of 788 instances, the clustering results are not stable for all three algorithms. Note that this result can be considered natural because these clustering algorithms are based on Euclidean distance measure which is only suitable for partition-based clustering problems, while Aggregation dataset contains data points distributed in contiguous regions and in different densities and sizes which are typically related to density-based clustering problems.

To sum up, on each of A1, A2, S1, S2, and R15 datasets where the clusters have apparent centroids and have similar numbers of data members compared to each other, our proposed POCS-based clustering algorithm and the K-Means algorithm share a similar performance and perform somewhat better than the FCM algorithm in terms of visual clustering results because the FCM algorithm sometimes still converges to sub-optimal solutions as can be seen from its results on A1 and A2 datasets in Fig. 4. Meanwhile, these algorithms are not suitable for working on density-based clustering problems such as Aggregation dataset.

In addition, the execution time is also considered as a comparison standard to assess the performance of those clustering algorithms. Table II summarizes the experimental results on execution times of different clustering methods. The execution speed of each algorithm is measured by executing the algorithm 10 times and deriving the mean value. As can be seen in Table II, the three algorithms can be roughly sorted according to the ascending execution times as follows: POCS-based, K-Means, and FCM.

\begin{table}[t]
\caption{Execution time comparison on various datasets (in seconds).}
\setlength\tabcolsep{0pt} 
\begin{tabular*}{\columnwidth}{@{\extracolsep{\fill}} lcccccc}

\toprule
      & \textbf{A1} & \textbf{A2} & \textbf{S1} & \textbf{S2} & \textbf{R15} & \textbf{Aggregation}  \\

\midrule
     \textbf{K-Means} & 0.09 & 0.30 & 0.09 & \textbf{0.09} & 0.04 & 0.03   \\
     \textbf{FCM} & 0.57 & 3.27 & 0.57 & 0.62 & 0.06 & 0.04   \\
     \textbf{POCS-based} & \textbf{0.08} & \textbf{0.20} & \textbf{0.08} & 0.11 & \textbf{0.03} & \textbf{0.02}   \\

\bottomrule
\end{tabular*}
\end{table}

\begin{table}[t]
\caption{Comparison in terms of mean and standard deviation of clustering error on various datasets.}
\setlength\tabcolsep{0pt} 
\begin{tabular*}{\columnwidth}{@{\extracolsep{\fill}} lccc}

\toprule
      & \textbf{K-Means} & \textbf{FCM} & \textbf{POCS-based} \\

\midrule
     A1  & 101.4 ± 7.1 & \textbf{88.8} ± 5.5 & 90.4 ± \textbf{4.9} \\
     A2  & 172.5 ± 10.7 & 175.8 ± 8.7 & \textbf{159.5} ± \textbf{8.6} \\
     S1  & 265.3 ± 44.9 & \textbf{198.9} ± 23.5 & 205.2 ± \textbf{21.3} \\
     S2  & 270.6 ± 29.8 & 233.3 ± \textbf{12.8} & \textbf{228.2} ± 13.3 \\
     R15 & 27.0 ± 6.4 & \textbf{16.7} ± 2.3 & 19.3 ± \textbf{2.1} \\
     Aggregation & 80.5 ± 2.1 & 81.8 ± 2.6 & \textbf{80.3} ± \textbf{1.8} \\

\bottomrule
\end{tabular*}
\end{table}

Clustering error is one of the most important measurements that is adopted to evaluate performance of clustering algorithms. The clustering error in our experiments is defined as:

\begin{equation}
E = \sum_{i=1}^{K} \sum_{j=1}^{N_i} ||c_i - x_{i,j} ||  \label{eq14}
\end{equation} where $K$ is the number of clusters, $N_i$, $c_i$, and $x_{i,j}$ are the number of data points, the final prototype, and the $j^{th}$ member data point of the $i^{th}$ cluster, respectively.

Table III summarizes the clustering error of different algorithms after convergence. The clustering error of each algorithm is computed by running the algorithm 20 times on a dataset and the mean and the standard deviation of the error are adopted as evaluation metrics. Note that all data points in each dataset are normalized to have values ranging from 0 to 1 for clustering error calculation. According to the results presented in Table III, the difference in clustering error among the examined algorithms is trivial. However, the proposed POCS-based clustering algorithm has shown a competitive clustering error when compared to that of the FCM algorithm. In addition, the POCS-based clustering algorithm provides a stable result at different running times when it consistently shows minimal dispersion of clustering error compared to that of the other clustering methods. This makes the proposed POCS-based clustering algorithm the most stable and robust algorithm among the rest.

As a result, the proposed POCS-based clustering algorithm possesses the fast execution speed of the K-Means algorithm while achieving the favorable clustering error as the FCM algorithm.

\section{Conclusions}

In this paper, a novel clustering technique based on the projection onto convex set (POCS) method, called POCS-based clustering algorithm, is presented. The proposed POCS-based clustering algorithm considers each data point as a convex set and projects the cluster prototypes to each of its constituent instances to compute the new prototypes. Based on the experimental results on various synthetic datasets, the proposed POCS-based algorithm has shown a superior performance compared to the K-Means algorithm in most cases and competitive enough with the FCM algorithm with marginal performance difference in terms of clustering error. Furthermore, the execution speed and simplicity are additional important advantages of the POCS-based clustering algorithm over the FCM clustering algorithm. The POCS-based algorithm converges much faster and can result in a more stable clustering output as compared to the K-Means and FCM clustering algorithms. In general, experimental results show that the proposed POCS-based algorithm can be considered as a promising tool for various data clustering tasks.

\end{document}